\documentclass[letterpaper]{article} 
\usepackage[preprint]{aaai2027}  
\usepackage[hyphens]{url}  
\usepackage{graphicx} 
\usepackage{natbib}  
\usepackage{caption} 
\usepackage{algorithm}
\usepackage{algorithmic}
\usepackage{amsmath}
\usepackage{amssymb}
\usepackage{booktabs}
\usepackage{multirow}
\usepackage{marvosym} 

\newtheorem{definition}{Definition}
\newtheorem{lemma}{Lemma}
\newtheorem{proposition}{Proposition}

\newcommand{\Ztw}{\mathbb{Z}_{12}}
\newcommand{\pc}{\operatorname{pc}}
\newcommand{\rev}{\operatorname{rev}}
\newcommand{\Forms}{\mathcal{F}}
\newcommand{\Lang}{\mathcal{L}}
\newcommand{\Viol}{\mathcal{V}}
\newcommand{\Hard}{\mathcal{H}}
\newcommand{\Repair}{\rho}
\newcommand{\Comp}{S}
\newcommand{\Ev}{e}

\newcommand{\ccom}{\textsuperscript{1}}
\newcommand{\tsinghua}{\textsuperscript{2}}
\let\defaultthefootnote\thefootnote
\newcommand{\corrmark}{%
  \gdef\thefootnote{\Letter}%
  \footnote{Corresponding author.}%
}

\title{%
\makebox[0pt][r]{\raisebox{-0.1mm}[0pt][0pt]{\raisebox{-0.4\height}{\includegraphics[height=1.4cm]{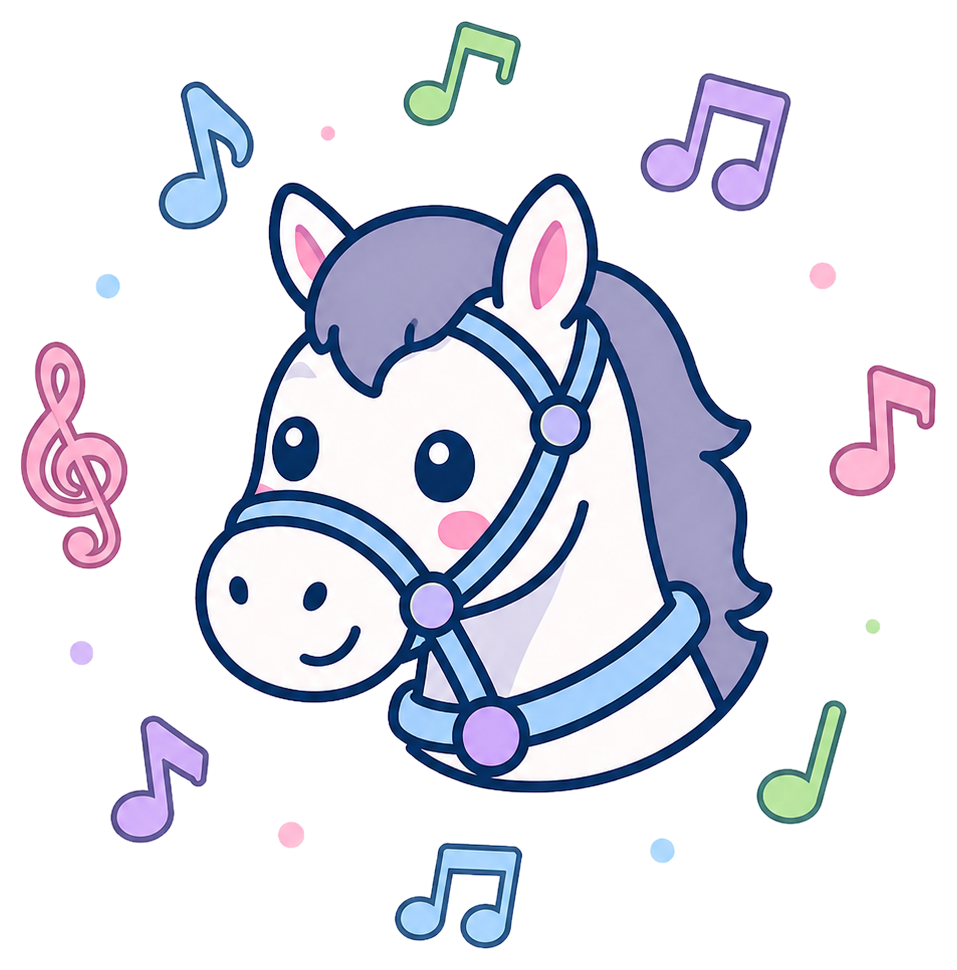}}}\hspace{0.5em}}%
\begin{tabular}[c]{@{}c@{}}
Verifier-Guided Twelve-Tone Composition:\\A Generate--Verify--Repair Harness for Symbolic Music Generation
\end{tabular}}

\author{Congren Dai\ccom\textsuperscript{,}\tsinghua,
Danni Zhao\ccom,
Enyang Liu\ccom,
Michael Ching Yam\ccom, \\
Zhancheng Guo\ccom\textsuperscript{,}\tsinghua, 
Siyi Gu\ccom,
Wentao Yang\ccom,
Bo Dai\ccom, 
Xiaobing Li\ccom, \textnormal{and} Maosong Sun\ccom\textsuperscript{,}\tsinghua\corrmark \vspace{5pt}}
\affiliations{\ccom\ Central Conservatory of Music \quad \tsinghua\ Tsinghua University \\
\vspace{2pt}
\small\texttt{congren.dai@mail.ccom.edu.cn \quad sms@tsinghua.edu.cn}}

\begin{document}

\maketitle
\global\let\thefootnote\defaultthefootnote

\begin{abstract}
Large language models can produce superficially legal twelve-tone scores that
collapse into degenerate textures. We introduce a neuro-symbolic harness that
wraps a language-model proposer in a \emph{generate--verify--repair--trace} loop with 
symbolic verification. The complete pipeline improves event-local
consistency without claiming whole-piece legality. Across $40$ controlled tasks
and four paired models, constraint-checked delivery rises from $13.3\%$
under raw generation to $48.1\%$ with the harness; it abstains on
the remaining $51.9\%$ runs.
The pass rate of a narrower collision and serialisation-consistency check rises
from $33.5\%$ to $58.3\%$, while degeneracy remains near $0.05$, including under 
adversarial prompting. A blinded evaluation by five experts also shows a
descriptive aggregate preference for harness candidates over raw
generation in adherence, perceived legality, coherence, and overall quality.
\end{abstract}


\section{Introduction}
\label{sec:intro}

Twelve-tone (serial) music, a compositional method in which an ordered
\emph{tone row} of pitch classes governs harmony and melody through transposition,
inversion, and retrograde \citep{schoenberg1950twelve,straus2016posttonal}, is
an attractive testbed for controllable generation: its core rules are crisp and
formal, yet musically interesting serial realisations often depend on global
organisation \citep{perle1991serial}. More generally, maintaining long-range
coherence is a central challenge in symbolic music generation
\citep{huang2019music}. This combination
exposes a failure mode
that is easy to miss when ``rule legality'' is the only metric. A model
instructed to prioritise legality can satisfy the letter of the rules with a \emph{degenerate}
score (a few sustained chords, collapsed voices, near-empty bars) that is
nominally compliant but musically empty. This behaviour is analogous to
\emph{specification gaming}: it satisfies the stated rule-based proxy while
defeating the intended musical objective. Reward-hacking work studies the
related, but not identical, case in which optimisation exploits a misspecified
objective or reward proxy
\citep{amodei2016concrete,pan2022reward,skalse2022reward}; here we elicit
shortcuts through adversarial prompts rather than training-time reward
optimisation.

We argue that a structural intervention can improve constraint satisfaction over prompting alone.
A deterministic verifier records hard-rule violations, and a row-aware repair
operator projects proposed material back onto the active row slice whenever
possible. The large language model (LLM) is retained where it is
genuinely useful (high-level planning and note-level proposals that drive
musical quality), while symbolic components control pitch-class assignment and
surface unresolved violations explicitly.
Unlike prompt-only self-correction, the symbolic layer evaluates executable
events rather than model-generated claims. Its trace records which row
positions survive repair, making dropped or modified material inspectable in
the delivered artefact.

This setting is a clean microcosm of a broader challenge in neuro-symbolic
generation: automatic proxies can be satisfied without producing useful
outputs, and process-level checks need to be distinguished from independent
checks on the delivered artefact. Twelve-tone
music sharpens the tension because legality is a crisp Boolean property, whereas
musical quality is global and is precisely what a legality-maximiser sacrifices.
Practically, the harness returns either a candidate passing a final constraint check or an explicit
failure with an inspectable record of retained, repaired, and dropped material.
Methodologically, separating process diagnostics from retained-artefact checks
offers a useful pattern for other structured generators, such as code or plans,
even though their domain predicates would need to be redesigned.
The target rule set is also unusually hard to enforce by decoding: it includes
cross-voice and long-range properties (each voice must track a scheduled
twelve-tone stream, and no two simultaneously sounding voices may share a pitch
class), so it cannot be
checked from a single token prefix. To study it we run six language models across
$40$ controlled tasks with length and texture sweeps, adversarial (red-team)
briefs, module ablations, a corpus-distance analysis against $20$ human serial
works, and a blind pairwise study with five experts.

\begin{figure*}[t]
\centering
\includegraphics[width=\textwidth]{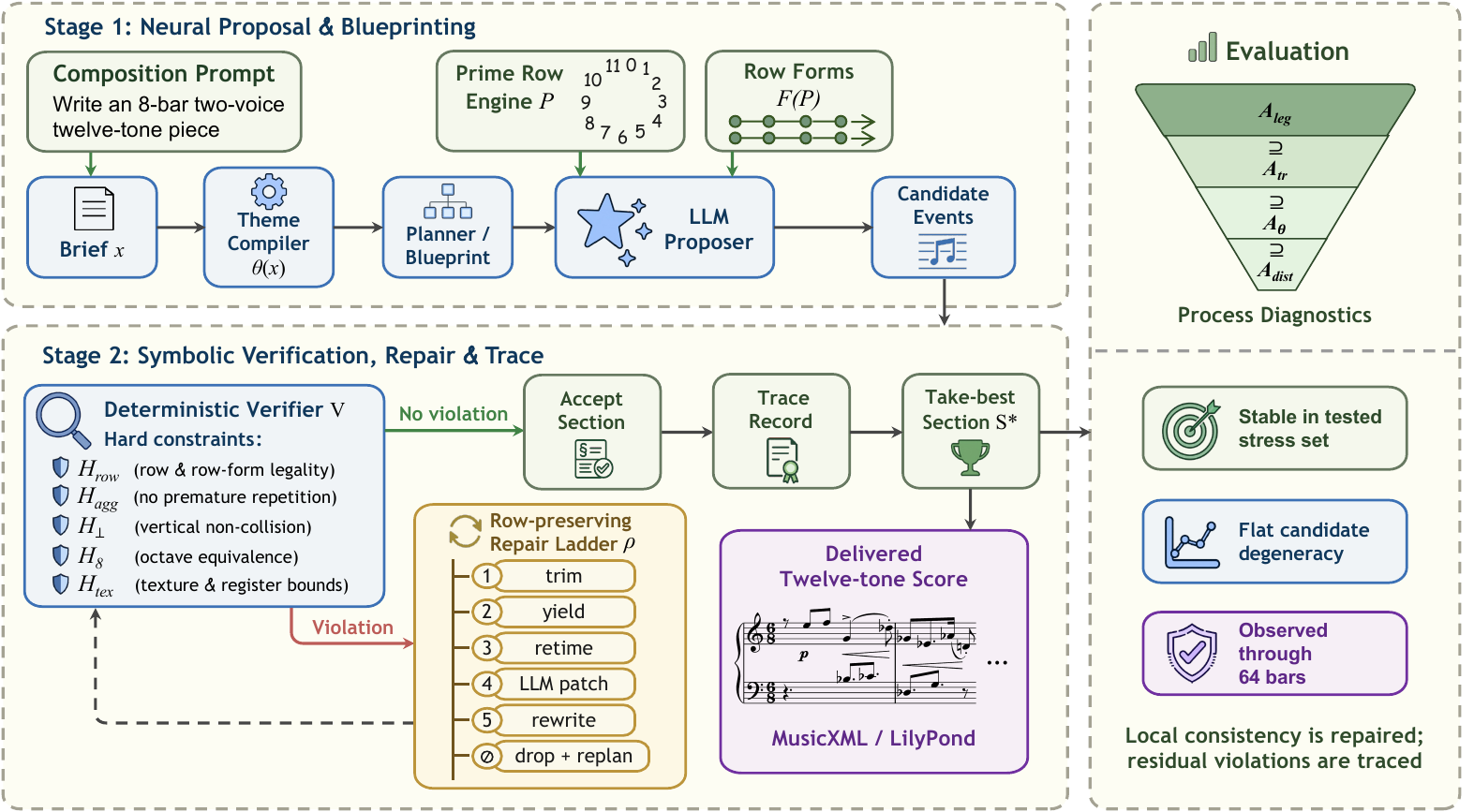}
\caption{Overview of the harness: an LLM proposer generates candidate events
(Stage~1), and a deterministic verifier with a row-aware repair ladder checks
or modifies them while recording a per-note trace (Stage~2). A final
retained-artefact constraint check gates release (Section~\ref{sec:tiers}).}
\label{fig:harness}
\end{figure*}

Concretely, we (1)~formalise serial composition as constrained sequence
generation over $\Ztw$: an order-$48$ transformation group, a core
event-consistency target defined by hard constraint predicates, and separate
process diagnostics and final-check decisions
(Section~\ref{sec:method}); (2)~introduce a \emph{generate--verify--repair--trace}
harness with a row-preserving repair operator and a generator/verifier
\emph{lockstep} invariant, yielding event-local row projection and monotone
best-attempt selection rather than unconditional whole-piece legality
(Section~\ref{sec:harness}); and (3)~evaluate, over the fixed task bank,
robustness to
degeneracy-inducing prompts, length and texture scaling, model capability, cost,
and expert preference.

\section{Related Work}
\label{sec:related}

\paragraph{Symbolic music generation.}
Prior systems address long-range structure \citep{huang2019music}, controlled
infilling \citep{thickstun2024anticipatory}, text-to-music attributes
\citep{lu2023musecoco}, and multi-track adversarial generation
\citep{dong2018musegan,muhamed2021tgan}. They are not organised around a
post-hoc verifier that reports domain-level rule violations. SerialGen instead
generates twelve-tone rows algorithmically \citep{feitosa2021serialgen}, without
an LLM proposer or verify--repair loop.

\paragraph{Constrained and grammar-constrained decoding.}
Prefix-constrained methods reject invalid next tokens using incremental parsers
or completion engines
\citep{scholak2021picard,poesia2022synchromesh,geng2023grammar}. Other approaches
use inference-time energies \citep{qin2022cold}, training-time logical losses
\citep{ahmed2023psemi}, or constraint programming followed by LM ranking
\citep{bonlarron2024cpnlp}. Our cross-voice, complete-row constraints instead
use an external stateful verifier paired with repair.

\paragraph{Verifier-guided generation and self-correction.}
Related approaches include process reward models
\citep{lightman2024verify}, generative verifiers
\citep{zhang2024genverifiers}, self-refinement \citep{madaan2023selfrefine}, and
feedback-based reflection \citep{shinn2023reflexion}. Unlike these
model-mediated approaches, our harness pairs deterministic symbolic checking
with explicit repair; Soft rules remain a one-shot, no-verifier baseline.

\paragraph{Reward hacking and specification gaming.}
Optimising an imperfect reward proxy can reduce performance under the true
objective \citep{amodei2016concrete,skalse2022reward,gao2023scaling}.
Examples include capability-threshold failures \citep{pan2022reward} and
reward-model over-optimisation \citep{gao2023scaling}; narrower rewards are not
generally unhackable \citep{skalse2022reward}. We test analogous prompted
shortcuts using delivered-score metrics rather than process diagnostics alone.

\section{Problem Formulation and Methodology}
\label{sec:method}

\subsection{Design Rationale}
\label{sec:rationale}

\paragraph{Twelve-tone music as a target.}
Twelve-tone composition suits generate--verify--repair because its serial
operations and aggregate structure are precisely defined
\citep{perle1991serial,babbitt1960invariants}. We operationalise them as
predicates over timed events, adding explicit row-form tags, aggregate
completion, and cross-voice non-collision checks
(Section~\ref{sec:lang}). A per-voice row cursor lets repair project a candidate
onto its tagged row slice; the LLM handles planning, voicing, and rhythm, while
the symbolic layer checks hard rules and records per-note traces.

\paragraph{Event representation.}
Internally, a composition is a multiset of timed events with MIDI-integer
pitches (Section~\ref{sec:lang}). The compact schema exposes pitch class
$m\bmod 12$, onset, duration, voice, row tag, and register directly to the
predicates and repair operators, avoiding notation-format parsing.

\paragraph{Notation and rendering.}
Composition, verification, and repair use only the canonical event record.
Our renderer deterministically exports the same retained events to LilyPond
and MusicXML. All blind-evaluation conditions share this implementation path, while
corpus-distance analysis uses the same monophonic pitch-line reduction
(Appendix~\ref{app:e5}).

\subsection{Pitch Classes and the Row-Form Group}
\label{sec:prelim}

Let $\Ztw=\{0,1,\dots,11\}$ be the set of pitch classes under octave
equivalence \citep{straus2016posttonal}. A \emph{prime row} (tone row) is an
ordered aggregate
$P=(P_0,\dots,P_{11})$, i.e.\ a bijection $\{0,\dots,11\}\to\Ztw$; every pitch
class occurs exactly once \citep{schoenberg1950twelve,perle1991serial}. Three
canonical operators act on rows:
\begin{align}
\text{transposition} \quad & T_n(x) = (x+n) \bmod 12, \\
\text{inversion} \quad & I(x) = (-x) \bmod 12, \\
\text{retrograde} \quad & R(x_0,\dots,x_{11}) = (x_{11},\dots,x_0).
\end{align}
$T_n$ and $I$ act pointwise on a row; $R$ reverses its time order. Writing
$\tilde P = T_{-P_0}(P)$ for the \emph{zeroed} prime (so $\tilde P_0=0$), the four
base forms, $P$ (prime), $I$ (inversion), $R$ (retrograde), and $RI$
(retrograde inversion), are
\begin{equation}
b_P=\tilde P,\;\; b_I = I\!\circ\!\tilde P,\;\; b_R=\rev(\tilde P),\;\; b_{RI}=\rev(I\!\circ\!\tilde P),
\end{equation}
and the \emph{row forms} are their transpositions
\begin{equation}
\Forms(P) = \{ f_{X,n} := T_n \circ b_X : X\in\{P,I,R,RI\},\ n\in\Ztw \}.
\end{equation}
These are the standard four row-form families
\citep{babbitt1960invariants,perle1991serial}. Since
$\langle T_n, I\rangle$ is the dihedral group $D_{12}$ of order $24$ and $R$
commutes with both pitch-class operations, retrograde contributes an
independent factor of $2$. The transformation operations therefore form
$D_{12}\times C_2$, of order $48$. Their action on a row produces at most $48$
distinct row forms, because symmetries may cause forms to coincide; hence
$\lvert\Forms(P)\rvert\leq 48$. Each
$f_{X,n}=(f_{X,n}[0],\dots,f_{X,n}[11])$ is itself an aggregate.

Two structural properties of $P$ are used by setup-time constraints. Following
the standard notion of derived rows \citep{babbitt1960invariants}, we call $P$ a
\emph{derived row} when its disjoint cells are related by standard serial
transformations. For reproducible evaluation, our implementation
operationalises this notion for cell sizes $c\in\{3,4,6\}$ ($c\mid 12$) by
requiring every disjoint length-$c$ block in the partition of $P$ to share the canonical
normalisation obtained by taking the minimum over
\{forward, retrograde, inverted, inverted-retrograde\}. Two forms
$f,g$ are \emph{hexachordally combinatorial} if their first hexachords partition
the aggregate: $\{f[0..5]\}\cap\{g[0..5]\}=\varnothing$ and
$\{f[0..5]\}\cup\{g[0..5]\}=\Ztw$
\citep{babbitt1960invariants}.

\subsection{Compositions, Traces, and Core Event Consistency}
\label{sec:lang}

An \emph{event} is a tuple
$\Ev=(\tau,\delta,\vec m,v,\phi,\sigma)$ with onset $\tau\in\mathbb{Q}_{\ge0}$,
duration $\delta>0$, MIDI pitches $\vec m\in\mathbb{Z}^k$ ($k\ge1$; $k>1$ is a
chord), a voice label $v$, a row-form tag $\phi\in\Forms(P)$, and an ordered
tuple of row indices $\sigma=(s_1,\ldots,s_r)$. Its \emph{pitch-class content} is
$\pc(\Ev)=\{m \bmod 12 : m\in\vec m\}$ and its \emph{sounding interval} is
$\iota(\Ev)=[\tau,\tau+\delta)$. A \emph{composition} is a finite multiset of
events $\Comp=\{\Ev_1,\dots,\Ev_N\}$; its restriction to voice $v$, written
$\Comp_v$, is the event stream of that voice ordered by onset.
For a retained candidate, $\widetilde{\mathbf p}_{\mathcal T}(\Ev)$ denotes the
ordered pitch-class tuple stored separately in the trace; unlike
$\pc(\Ev)$, it is not defined by reducing $\vec m$ modulo $12$.

\paragraph{Hard constraints.}
For fixed prime row $P$ and blueprint $B$, each hard rule is a predicate
$H_{P,B}:\,\Comp\mapsto\{\top,\bot\}$. The formal target set $\Hard$ is
summarised in Table~\ref{tab:hard} (Appendix~\ref{app:metrics}); the online
harness records attempted enforcement, while the final release check evaluates
a broader subset directly on retained artefacts. The principal
target predicates are:
\begin{itemize}
\item \textbf{Tagged row consistency} ($H_{\mathrm{row}}$): each retained
event satisfies $\pc(\Ev)=\{f_{\phi}[s]:s\in\sigma\}$ for its tagged
$(\phi,\sigma)$, and the ordered tags are monotone in the scheduled row-form stream. Because
an unrepaired proposal may be dropped while its cursor advances, the retained
line can be a tag-preserving subsequence of that stream; it need not be a
complete concatenation of forms.
\item \textbf{Aggregate / no premature repetition} ($H_{\mathrm{agg}}$):
within every uninterrupted retained row cycle, no pitch class recurs before
the cycle advances, except when the online state admits a local
(oscillation/tremolo) window. In the implementation this window is inferred
from recent cursor history rather than pre-authorised by a separate blueprint
field. Aggregate \emph{completion} is reported separately because dropped
events can leave a cycle incomplete.
\item \textbf{Vertical non-collision} ($H_{\perp}$): for events
$\Ev,\Ev'$ in distinct voices with overlapping sounding intervals
($\lvert\iota(\Ev)\cap\iota(\Ev')\rvert>\epsilon$),
$\pc(\Ev)\cap\pc(\Ev')=\varnothing$; the positive tolerance $\epsilon$ guards
against floating-point boundary contact. The process verifier uses
$\epsilon=10^{-6}$ beats; the independent comparison checker uses the more
conservative $10^{-3}$-beat threshold reported with that metric.
\item \textbf{Octave equivalence} ($H_{8}$): the independently stored trace
tuple agrees with the retained MIDI pitches,
\[
\begin{aligned}
H_8(\Ev,\mathcal T)&\iff
\widetilde{\mathbf p}_{\mathcal T}(\Ev)\\
&=(m_1\bmod12,\ldots,m_k\bmod12).
\end{aligned}
\]
\item \textbf{No global rest / voice independence} ($H_{\mathrm{tex}}$): in
sections with at least two active voices, at every time in the active span at
least one voice sounds; and the fraction of noteheads participating in any
simultaneous higher/lower-voice register inversion is at most $0.1$.
\end{itemize}
Setup-time predicates ($H_{\mathrm{seg}}$, $H_{\mathrm{der}}$,
$H_{\mathrm{comb}}$) check segmentation length consistency, derived-row identity,
and hexachordal combinatoriality against $P$ and the blueprint.

\begin{definition}[Core event-consistency language]
The \emph{core event-consistency language} for $P$ and $B$ is
$\Lang(P,B) = \{\Comp : \forall H_{P,B}\in\Hard,\ H_{P,B}(\Comp)=\top\}$.
\end{definition}
This target language does not by itself require nonempty output, realisation of
every requested voice, or completion of every aggregate; those coverage
properties are reported separately. We therefore do not use membership in
$\Lang(P,B)$ as a synonym for complete twelve-tone legality.

\begin{algorithm}[t]
\caption{Generate--verify--repair candidate construction for one section: release occurs only after the final constraint check.}
\label{alg:gvr}
\textbf{Input}: section spec, voice states $\{u_v\}$, row engine on $P$\\
\textbf{Parameter}: max attempts $K$ ($K=3$ with replan; $K=1$ without)\\
\textbf{Output}: retained candidate events (release follows the final check)
\begin{algorithmic}[1]
\STATE $\Comp^\star \leftarrow \varnothing$;\; $m^\star \leftarrow \infty$
\STATE $\{u_v^0\}\leftarrow\textsc{Snapshot}(\{u_v\})$
\FOR{$\mathrm{att} = 1$ \TO $K$}
  \STATE $\{u_v\}\leftarrow\textsc{Clone}(\{u_v^0\})$ \COMMENT{no cursor state leaks across attempts}
  \STATE propose candidate events per voice (LLM or deterministic), concurrently
  \STATE sort candidates by $(\tau, v)$; assign event ids \COMMENT{determinism}
  \FOR{each candidate $\Ev$ in order}
    \STATE $\Ev' \leftarrow \Repair^\star(\Ev \mid \text{accepted}, u_v)$ \COMMENT{bounded escalation}
    \IF{$\Ev' \neq \bot$}
      \STATE accept $\Ev'$
    \ENDIF
    \STATE $\textsc{Advance}(u_v)$ \COMMENT{cursor advances even if dropped}
  \ENDFOR
  \STATE evaluate section-level and multi-voice hard rules; let $\Comp_{\mathrm{att}}$ be the result
  \IF{$\lvert\widehat{\Viol}_{\mathrm{proc}}(\Comp_{\mathrm{att}})\rvert < m^\star$}
    \STATE $\Comp^\star \leftarrow \Comp_{\mathrm{att}}$;\; $m^\star \leftarrow \lvert\widehat{\Viol}_{\mathrm{proc}}(\Comp_{\mathrm{att}})\rvert$
  \ENDIF
  \IF{$m^\star = 0$} \STATE \textbf{break} \ENDIF
  \STATE \textsc{Replan}: reassign row forms to minimise simulated collisions (guarded)
\ENDFOR
\STATE \textbf{return candidate} $\Comp^\star$
\end{algorithmic}
\end{algorithm}

The implementation reconstructs the per-voice states from the events accepted
before this section at the start of every attempt. Cursor advances made by a
failed attempt therefore do not affect later attempts; only the operative
section specification may change through replanning.
The algorithmic replan may alter the operative section and row-form schedule
within an attempt. The retained result currently stores the initial blueprint
rather than a fully materialised history of those local replans. Consequently,
agreement with the stored schedule is reported separately as a provenance
check and is not implied by a process pass.

An ideal final-score verifier $\Viol$ maps a composition to the multiset of hard violations,
$\Viol(\Comp)=\{(H,w): H\in\Hard,\ w\text{ witnesses }\neg H(\Comp)\}$, so
$\Comp\in\Lang(P,B) \iff \Viol(\Comp)=\varnothing$. Soft (aesthetic) rules are
collected separately as warnings and never affect membership in $\Lang$.
The implemented process diagnostic instead reads violations accumulated during
online generation; write this list as $\widehat{\Viol}_{\mathrm{proc}}$. It is
not a fresh evaluation of every predicate on the final normalised candidate.

\paragraph{Traces.}
Every accepted event $\Ev$ carries a \emph{certificate}
$c_{\mathcal T}(\Ev)=(\tau,\delta,\vec m,
\widetilde{\mathbf p}_{\mathcal T},v,q,\phi,\sigma,
\text{status},\text{owner})$, where $q$ is the section identifier. Thus MIDI
pitches and claimed pitch classes are separately serialised rather than
identified by definition. The implementation additionally stores event ids,
repair history, and acceptance flags. The trace
$\mathcal{T}(\Comp)=\{c_{\mathcal T}(\Ev):\Ev\in\Comp\}$ is a per-event
provenance log; it is
used by higher process-acceptance tiers (Section~\ref{sec:tiers}) to test whether
a claimed pitch class is derivable from a recorded row position. It supports
event-level inspectability, not complete planning provenance or an independent
certificate of the entire retained candidate.

\subsection{The Generate--Verify--Repair Harness}
\label{sec:harness}

The harness maps a natural-language brief $x$ to either a candidate passing the
specified release check or an explicit failure status. It first constructs a candidate $\Comp^\star$,
with the design goal of reducing
$\lvert\widehat{\Viol}_{\mathrm{proc}}(\Comp^\star)\rvert$ to zero within a
bounded attempt budget. A theme compiler turns $x$ into a
structured specification $\theta(x)$ (bar counts, voice counts, textures, derived
/ combinatorial requests, etc.), which a planner expands into a blueprint of
sections; the prime row $P$ is generated deterministically from the compiled
specification once per task and is then held fixed across models, conditions,
and the three runs. Within each section the harness runs Algorithm~\ref{alg:gvr};
after assembly, the final constraint check of Section~\ref{sec:tiers} gates release.

\paragraph{Composer-facing control.}
The harness also accepts hand-authored blueprints through the same
verify--repair procedure. Composers can specify $P$, up to four voices, and
per-voice row forms and schedules, register $[\ell_v,h_v]$, rhythmic density,
trichordal/tetrachordal/hexachordal segmentation, optional hexachordal
combinatoriality, and staggered entries. These are executable inputs: repair
enforces registers, although local replanning may revise schedules. This
supports direct control over an individual serial practice, but is not
evaluated; the benchmark uses only blueprints automatically synthesised from
briefs.

\paragraph{Row-preserving repair.}
The repair operator is an ordered escalation ladder
$\Repair=(\Repair_1,\dots,\Repair_5)$ applied to a flagged event;
$\Repair^\star(\Ev)$ returns the first $\Repair_j(\Ev)$ that clears the local
violation, or $\bot$ (drop the event and defer its slot to a section replan) if
none do. The first three rungs target vertical
collisions by editing \emph{only timing}:
\begin{equation}
\Repair_j(\Ev) = (\tau',\delta',\vec m,v,\phi,\sigma), \quad j\in\{1,2,3\},
\end{equation}
i.e.\ $\Repair_{1\text{--}3}$ change $(\tau,\delta)$ but leave
$(\vec m,\phi,\sigma)$ fixed: rung~1 trims the current note's tail, rung~2 yields
by trimming an overlapping sustained note in another voice, rung~3 retimes the
note inside its own free window. The last two rungs are LLM patches checked
again after register clamping (rung~4 edits the single event and rung~5 rewrites
its row segment), after which the event is dropped and the section is flagged
for replan. The following invariance applies only to the deterministic timing
rungs:

\begin{lemma}[Repair invariance]
\label{lem:rowpres}
For $j\in\{1,2,3\}$, $\pc(\Repair_j(\Ev))=\pc(\Ev)$ and the tagged
$(\phi,\sigma)$ are unchanged. Hence the deterministic rungs preserve
$H_{\mathrm{row}}$ and $H_{\mathrm{agg}}$ and do not alter MIDI pitch content.
$H_8$ is evaluated after the repaired event is serialised into the trace.
Since the rungs only \emph{shorten} or \emph{move into free space}, they
introduce no new vertical
collision or intra-voice overlap. Rung~2 additionally \emph{shortens} the
overlapping sustained note in the \emph{other} voice; being again only a
truncation, it creates no new collision and leaves that voice's pitch classes
and row tags intact, so the invariance extends to every event the rung edits.
\end{lemma}

\paragraph{Lockstep invariant.}
Generation and verification maintain a shared row cursor per voice. Even when
$\Repair^\star(\Ev)=\bot$, the \textsc{Advance} step of Algorithm~\ref{alg:gvr}
still increments the cursor, so the verifier's expected row index never lags the
generator's.

\begin{proposition}[Cursor lockstep]
\label{prop:lockstep}
At every step, for each voice $v$ the verifier-side next index equals the
generator-side next index. Consequently a dropped event costs exactly one rest
at its slot and does not desynchronise later tags. More generally, a dropped
multi-index candidate omits its tagged row slice. The retained stream is
therefore a tag-preserving subsequence, not necessarily a complete row
statement.
\end{proposition}

\paragraph{Bounded best-attempt selection.}
Let $\Comp^{(1)},\dots,\Comp^{(K)}$ be the per-attempt results and
$\Comp^\star=\arg\min_k \lvert\widehat{\Viol}_{\mathrm{proc}}(\Comp^{(k)})\rvert$ the retained candidate
(``take-best'').

\begin{proposition}[Projection and selection]
\label{prop:legal}
(i) Within a fixed attempt and relative to its operative local section state,
candidate construction, tag checking, and the deterministic repair rungs
preserve each retained event's expected row pitch class and octave-equivalent
tag; later events remain aligned after a drop by
Proposition~\ref{prop:lockstep}. This is an event-local invariant and does not
imply that every requested aggregate is complete or that the event schedule
matches the initial blueprint retained in the artefact. (ii) If
$M_K=\min_{1\leq k\leq K}\lvert\widehat{\Viol}_{\mathrm{proc}}(\Comp^{(k)})\rvert$, then
\begin{equation*}
M_{K+1}\leq M_K,\qquad
\lvert\widehat{\Viol}_{\mathrm{proc}}(\Comp^\star)\rvert=M_K,
\end{equation*}
so retaining an additional attempt cannot worsen the best observed violation
count. (iii) No existence claim follows from the bounded search: the retained
candidate can contain violations and therefore fail the final release gate.
\end{proposition}

\subsection{Process Diagnostics and Final Release Check}
\label{sec:tiers}

For harness runs, we retain a chain of increasingly strict \emph{process}
acceptance predicates over a composition $\Comp$ for brief $x$, supporting
transparent stage-wise diagnostic reporting:
\begin{align}
A_{\mathrm{leg}}(\Comp) &= \bigl[\widehat{\Viol}_{\mathrm{proc}}(\Comp)=\varnothing\bigr],\\
A_{\mathrm{tr}}(\Comp) &= A_{\mathrm{leg}}(\Comp)\wedge \bigl[\lvert\mathcal{T}(\Comp)\rvert=\lvert\Comp\rvert\bigr],\\
A_{\theta}(\Comp) &= A_{\mathrm{tr}}(\Comp)\wedge \Theta\bigl(\Comp,\theta(x)\bigr),\\
A_{\mathrm{dist}}(\Comp) &= A_{\theta}(\Comp)\wedge C_{\mathrm{dist}}(\Comp),
\end{align}
Here $\Theta$ checks bar/voice counts and requested techniques, while
$C_{\mathrm{dist}}$ requires attacks in at least $75\%$ of each active voice's
bars. The diagnostics are nested:
\begin{equation}
A_{\mathrm{dist}} \Rightarrow A_{\theta} \Rightarrow A_{\mathrm{tr}} \Rightarrow A_{\mathrm{leg}}.
\end{equation}
These tiers reuse process records and are diagnostics, not release decisions.
After score assembly, the separately implemented final constraint check
$A_{\mathrm{check}}$ (Appendix~\ref{app:metrics}) rescans the retained candidate.
The output contract is
\[
\textsc{Compose}(x)=
\begin{cases}\Comp^\star,&A_{\mathrm{check}}(\Comp^\star)=1,\\
\textsc{Failure},&\text{otherwise}.
\end{cases}
\]
The gate does not trigger further generation; failed candidates remain stored
only for analysis. Stored-blueprint agreement is reported separately because
operative replans are not fully materialised.

\begin{figure*}[t]
\centering
\includegraphics[width=\textwidth]{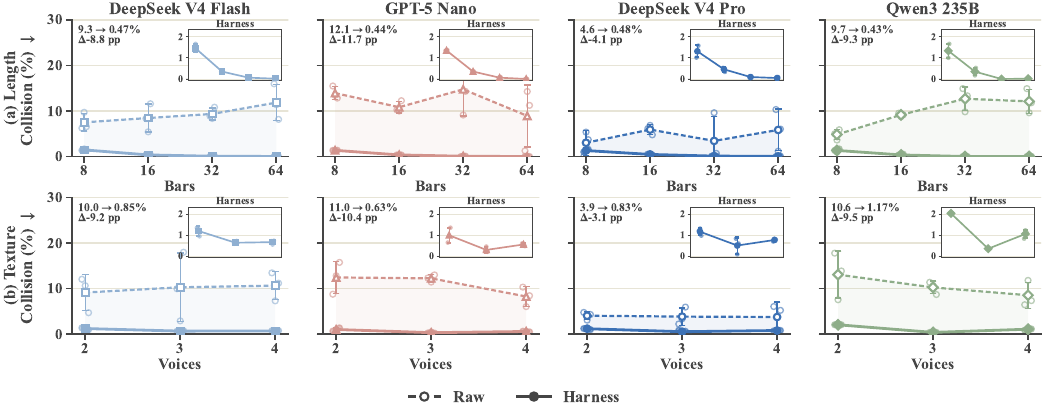}
\caption{Cross-voice collision rates by model. Rows vary length (top) and
texture (bottom); columns separate models. Curves show three-run means
$\pm$ one standard deviation, faint points show runs, and insets magnify
harness rates. Raw rates are conditional on parseability; one voice is omitted
because no cross-voice denominator exists.}
\label{fig:reliability}
\vspace{-5pt}
\end{figure*}

\section{Experimental Design}
\label{sec:experiments}

\paragraph{Tasks.}
We use $40$ single-factor tasks: a $20$-task length sweep
($8$--$64$ bars, two voices) and a $20$-task texture sweep
($1$--$4$ voices, eight bars), sharing a two-voice/eight-bar anchor. Each task
contains one mild writing concern. Full briefs and inventories appear in
Appendix~\ref{app:briefs}.

\paragraph{Models.}
DeepSeek V4 Pro/Flash, GPT-5 Nano, and Qwen3 235B-A22B (Qwen3 235B) run both raw
and in-harness. The full model inventory appears in
Appendix~\ref{app:tasks}.

\paragraph{Expert evaluation.}
Five experts compare $24$ anonymised Qwen3 235B pairs
($8$ tasks $\times$ three contrasts).

\paragraph{Experimental setup.}
The four paired models share one API and decoding configuration
(Appendix~\ref{app:inference}). A task's prime row is
fixed across models, conditions, and runs; seeds control only LLM sampling.
After averaging runs within task, we report point estimates and
$2000$-resample nonparametric family-clustered bootstrap
\citep{efron1979bootstrap} confidence intervals (CIs). Each draw samples the
ten prompt families with their four tasks.
The ablation study comprises the deterministic backbone, raw generation, Self-Refine,
Soft rules, and the full harness.

\paragraph{Metrics.}
We report process diagnostics, constraint-wise marginal pass rates, constraint-checked
delivery yield, and a
narrower, separately implemented collision and serialisation-consistency check.
Other outcomes are corpus squared maximum mean discrepancy
(MMD$^2$) \citep{gretton2012mmd} with a Gaussian radial basis function (RBF)
kernel and Kullback--Leibler (KL) divergence \citep{kullback1951}
(Appendix~\ref{app:e5}), degeneracy (Appendix~\ref{app:metrics}), cost
(Appendix~\ref{app:figs}), and expert preference
(Appendix~\ref{app:expert}). Red-team analysis combines
degeneracy, retained material, and the independent check rather than relying
on process fields.

\section{Results and Analysis}
\label{sec:results}

\begin{table}[t]
\centering
\caption{Ablation and prompt-only deltas versus matching full-harness
baselines. The top block is deterministic; lower blocks use Qwen3 235B.
Values cover $40$ paired tasks; ``accept'' is a process outcome, not an
independent legality measure.}
\label{tab:ablation}
\setlength{\tabcolsep}{4pt}
\resizebox{\columnwidth}{!}{%
\begin{tabular}{@{}lccc@{}}
\toprule
Condition & $\Delta$ process accept & $\Delta$ ind.\ core & $\Delta$ deg. \\
\midrule
\multicolumn{4}{@{}l}{\emph{Deterministic backbone (vs.\ full harness)}}\\
\;No repair            & $-$0.192 & $+$0.050 & $-$0.006 \\
\;No repair + no replan & $-$0.192 & $+$0.025 & $-$0.004 \\
\;No replan             & $-$0.025 & $+$0.025 & $+$0.001 \\
\midrule
\multicolumn{4}{@{}l}{\emph{LLM in the loop (vs.\ full harness)}}\\
\;No skills             & $+$0.017 & $-$0.042 & $+$0.000 \\
\;Fixed row             & $-$0.008 & $-$0.092 & $+$0.001 \\
\midrule
\multicolumn{4}{@{}l}{\emph{Prompt-only baselines (vs.\ full harness)}}\\
\;Self-refine           & $-$0.183 & $-$0.333 & $+$0.241 \\
\;Soft rules            & $-$0.237 & $-$0.400 & $+$0.208 \\
\bottomrule
\end{tabular}
}
\vspace{-15pt}
\end{table}

\subsection{Length and Texture Scaling}
\label{sec:reliability}

\begin{figure*}[!t]
\centering
\includegraphics[width=\textwidth]{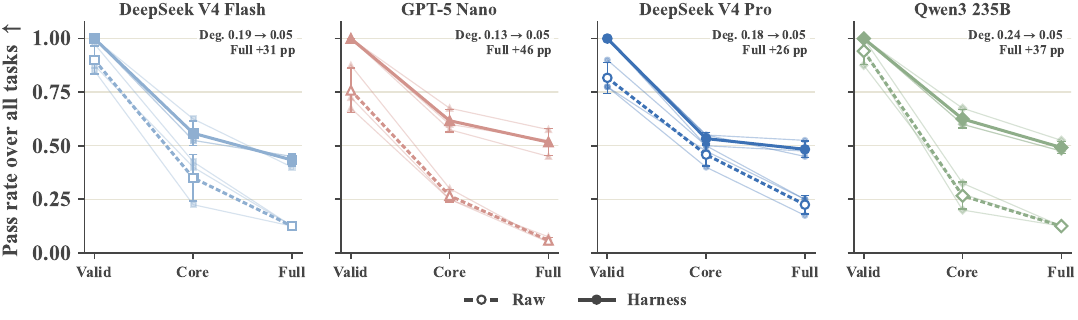}
\caption{Raw-to-harness reliability ladder by model: parseable output
(\emph{Valid}), independent collision and serialisation pass (\emph{Core}), and
the schedule-relaxed release check (\emph{Full}). Faint lines show runs;
bold lines show means $\pm$ one standard deviation. Annotations give degeneracy (Deg.)
and the all-task Full-rate gain.}
\label{fig:ladder}
\vspace{-5pt}
\end{figure*}

Figure~\ref{fig:reliability} separates model and seed variation in the local
collision diagnostic. Across plotted settings, harness three-run means remain
at or below $2.05\%$, whereas raw three-run means, conditional on parseability,
reach $14.76\%$. Raw length trends are model-dependent; harness rates generally
fall toward zero as length grows. Rates are opportunity-normalised
(Appendix~\ref{app:metrics}).
Figure~\ref{fig:ladder} complements this local metric with an end-to-end
filtering sequence over all $480$ attempted tasks. From raw to harness, parseable output
(\emph{Valid}) rises from $410/480$ ($85.4\%$) to $480/480$; the independent
collision and serialisation check (\emph{Core}) rises from $33.5\%$ to $58.3\%$;
and constraint-checked delivery (\emph{Full}) rises from $64/480$ ($13.3\%$) to $231/480$
($48.1\%$). Every model improves at Core and Full. The harness abstains on the
remaining $249$ candidates rather than releasing failed scores. Full rates and
denominator caveats appear in Appendix~\ref{app:metrics}; red-team parseability
appears in Appendix~\ref{app:redteam-full}.
This primary release check is schedule-relaxed: $184/480$ candidates disagree
with the initially stored schedule, including $61$ that pass every other check.
Requiring stored-blueprint agreement reduces delivery to $170/480$ ($35.4\%$).

\paragraph{Constraint-wise diagnosis.}
Table~\ref{tab:ablation} shows that repair raises process acceptance but not the
narrow candidate check. Without repair, candidates contain $11.2$ fewer
events ($0.71$ fewer notes/bar) on average, consistent with reduced collision
exposure; coverage changes by only $-0.001$. Thus, the $+0.050$ narrow-check
delta does not establish that repair harms candidate consistency. More details
appear in Appendix~\ref{app:ablation-full}. At the full-system level, marginal
rates show that the harness passes row-form/order ($99.4\%$), no
premature repetition ($99.8\%$), aggregate completion ($95.6\%$), and the
voice-crossing bound ($96.0\%$) in nearly all candidates. The remaining
bottlenecks are vertical non-collision ($58.3\%$) and the combined texture check
($74.6\%$). Vertical non-collision determines the equal-valued narrow
collision and serialisation-consistency conjunction because its other
components pass for every harness candidate. Thus, the $48.1\%$ delivery yield is a conjunction across
heterogeneous checks, not typical per-constraint performance
(Appendix~\ref{app:metrics}, Table~\ref{tab:constraint-marginals}).

\subsection{Robustness to Degeneracy-Inducing Prompts}
\label{sec:reward-hacking}

Tables~\ref{tab:redteam-raw} and~\ref{tab:redteam-harness}
(Appendix~\ref{app:redteam-full}) report the exploratory end-to-end prompt
stress test on Qwen3 235B. Because attack suffixes also pass through
the theme compiler, they can change the executable specification rather than
attacking a fixed proposer alone. Harness degeneracy stays near $0.05$ and its
collision and serialisation-consistency pass rate remains at $0.55$--$0.65$, but
the \emph{Long chords} variant eliminates theme coverage.

\subsection{Capability--Scaffolding Interaction and Cost}
\label{sec:democratisation}

Across four models, in-harness generation raises the independent collision and
serialisation pass rate to $0.53$--$0.63$, constraint-checked delivery to
$0.43$--$0.52$, and lowers degeneracy to $0.047$--$0.053$
(Table~\ref{tab:ladder-full}). These descriptive gains apply only to the tested
models and are costly: on Qwen3 235B, the harness uses $638$ calls, $823$k
tokens, and $841$ seconds per run, versus one call, $5.3$k tokens, and $77$
seconds for raw generation (Table~\ref{tab:cost}).

\begin{table}[t]
\centering
\caption{Mean per-run cost on Qwen3 235B
($n\approx110\text{--}120$). Tokens are provider-reported total usage;
wall-clock covers end-to-end execution.}
\label{tab:cost}
\setlength{\tabcolsep}{3pt}
\begin{tabular}{@{}lccc@{}}
\toprule
Condition & LLM calls & Tokens & Wall-clock (s) \\
\midrule
Blueprint only            & 1.0   & 2{,}433   & 24.0 \\
Raw LLM                   & 1.0   & 5{,}257   & 76.7 \\
Soft rules                & 1.0   & 5{,}117   & 62.8 \\
Self-refine               & 5.0   & 23{,}850  & 223.9 \\
Full harness, no skills   & 627.3 & 234{,}845 & 814.5 \\
Full harness              & 638.0 & 823{,}482 & 841.0 \\
\bottomrule
\end{tabular}
\vspace{-10pt}
\end{table}

\subsection{Musical Quality}
\label{sec:corpus}

Conditional on parseability, mean degeneracy falls for every model from
$0.132$--$0.244$ raw to $0.047$--$0.053$ in-harness
(Figure~\ref{fig:ladder}; Table~\ref{tab:ladder-full}). Because this diagnostic
captures structural collapse rather than musical quality,
Figure~\ref{fig:sample-score} provides a representative score.

\paragraph{Corpus distribution analysis.}
Table~\ref{tab:corpus} compares the retained harness candidates with a $20$-piece human
corpus using pitch-distribution features (Appendix~\ref{app:e5}). Harness
candidates are closer to the strict-serial than the free-atonal subset under both
MMD$^2$ and KL.

\paragraph{Task-completion analysis.}
Table~\ref{tab:task-completion} captures properties omitted by the corpus
features. Every harness candidate is parseable, covers $97.3\%$ of requested
voice--bar cells, and completes aggregates in $95.6\%$ of all runs; parseable raw
candidates cover only $63.5\%$ of voice--bar cells.

\begin{table}[t]
\centering
\caption{Corpus distribution distance for $480$ retained harness candidates. One RBF
bandwidth is shared. KL is interval-class / directed-interval / pitch-class,
generated$\,\|\,$corpus.}
\label{tab:corpus}
\normalsize
\setlength{\tabcolsep}{4pt}
\begin{tabular}{@{}lccc@{}}
\toprule
Corpus subset & $n$ & MMD$^2$ & KL (ic / di / pc) \\
\midrule
all           & 20 & 0.330 & 0.073 / 0.074 / 0.009 \\
strict-serial & 9  & 0.278 & 0.041 / 0.042 / 0.004 \\
free-atonal   & 11 & 0.419 & 0.145 / 0.155 / 0.024 \\
\bottomrule
\end{tabular}
\end{table}

\begin{table}[t]
\centering
\caption{Raw vs.\ harness candidate completion over $480$ paired cells. Values are
three-run mean $\pm$ standard deviation pooled over four models. Parseability and aggregate
completion use the intention-to-treat denominator; coverage and voice
realisation are conditional on parseable candidates.}
\label{tab:task-completion}
\setlength{\tabcolsep}{5pt}
\resizebox{\columnwidth}{!}{%
\begin{tabular}{@{}lcc@{}}
\toprule
Metric & Raw & Harness \\
\midrule
Parseable-candidate rate & $0.854\pm0.055$ & $1.000\pm0.000$ \\
Active-bar coverage & $0.635\pm0.020$ & $0.973\pm0.001$ \\
Voice realisation & $0.995\pm0.005$ & $1.000\pm0.000$ \\
Aggregate-completion pass & $0.673\pm0.046$ & $0.956\pm0.006$ \\
\bottomrule
\end{tabular}
}
\vspace{-5pt}
\end{table}

\subsection{Expert Blind Evaluation}
\label{sec:expert}

Five experts evaluate $24$ anonymised retained-candidate pairs. The full harness receives
$82$--$90\%$ of choices over raw generation for adherence, perceived legality,
coherence, and overall quality, and $70\%$ for style. Against the ablations,
full-harness win rates span $55$--$75\%$ without skills and $57$--$75\%$ without
repair. Task-clustered intervals and low inter-rater agreement qualify these
aggregate preferences (Appendix~\ref{app:expert}).

\section{Conclusion and Limitations}
\label{sec:conclusion}

We introduce a generate--verify--repair harness for constrained twelve-tone
composition. It raises schedule-relaxed, constraint-checked delivery from
$13.3\%$ to $48.1\%$ across four paired models, but abstains on $51.9\%$ of
runs; on Qwen3 235B, it averages $638$ calls and $823$k tokens per run. Its
trace supports event-level inspectability, not complete planning provenance,
and stored-blueprint agreement lowers delivery to $35.4\%$. The contribution is
therefore improved selective delivery and explicit failure within the tested
task bank, not production-ready
or reliably complete composition.

\FloatBarrier
\section*{Acknowledgements}
This work is supported by the Advanced Discipline Construction Project of Beijing Universities, the Special Programme of National Natural Science Foundation of China (Grant No. T2341003), and the Major Programme of National Social Science Fund of China (Grant No. 21ZD19).

\setlength{\bibsep}{1.7pt}
\bibliography{aaai2027}

\newpage
\appendix

\section{Corpus-Distance Features and Metrics}
\label{app:e5}

This appendix gives the full feature schema and distance definitions summarised
in Section~\ref{sec:corpus}. Everything is computed from monophonic pitch lines so
that generated candidates and the human corpus are compared on identical footing.

\paragraph{Pitch lines.}
Human scores are stored as MusicXML and parsed once with \texttt{music21}.
Each staff/part is split by explicit MusicXML voice; without voice elements, it
forms one line. Notes are ordered by measure and within-measure offset, rests
are omitted, chords retain only their highest MIDI pitch, and lines shorter
than two notes are discarded, yielding monophonic lines
$\mathcal{P}=\{\ell\}$. Generated records undergo the same onset ordering,
chord reduction, and length filter.

\paragraph{Feature vector.}
For a line $\ell=(p_1,\dots,p_T)$ we accumulate, over all lines of a piece, the
pitch classes $p_t \bmod 12$ and the nonzero directed intervals
$d_t = (p_{t+1}-p_t)\bmod 12$, with interval class $\mathrm{ic}(d)=\min(d,12-d)$.
Normalising each histogram to a probability vector yields three distribution
blocks: interval class $\mathbf{c}\in\Delta^{5}$ (dims $1$--$6$), directed
interval $\mathbf{d}\in\Delta^{10}$ (dims $1$--$11$), and pitch class
$\mathbf{h}\in\Delta^{11}$ (dims $0$--$11$). Four melodic scalars are appended:
the normalised Shannon entropy of $\mathbf{d}$, the fraction of the $11$ directed
intervals that occur, the \emph{leap ratio} (fraction of absolute steps
$\lvert p_{t+1}-p_t\rvert \ge 8$), and the mean absolute interval divided by $12$
(capped at $1$). Concatenation gives a fixed
$\phi(\cdot)\in\mathbb{R}^{33}$ ($6+11+12+4$).

\paragraph{Squared MMD.}
Let $X=\{x_i\}_{i=1}^{n}$ (corpus) and $Y=\{y_j\}_{j=1}^{m}$ (generated) be the
feature vectors. With the RBF kernel
$k(a,b)=\exp(-\gamma\lVert a-b\rVert^2)$ we report the
unbiased estimator of squared MMD \citep{gretton2012mmd}
\begin{equation*}
\begin{aligned}
\widehat{\mathrm{MMD}}_u^2 ={}&
\frac{\textstyle\sum_{i\ne j}k(x_i,x_j)}{n(n-1)}
+ \frac{\textstyle\sum_{i\ne j}k(y_i,y_j)}{m(m-1)} \\
&{}- \frac{2}{nm}\,\textstyle\sum_{i,j}k(x_i,y_j),
\end{aligned}
\end{equation*}
where $\gamma=1/\tilde{s}$ and $\tilde{s}$ is the median squared distance over
distinct unordered pairs in the fixed pool containing the full $20$-piece
corpus and all $480$ harness candidates, using the median-heuristic convention
analysed by \citet{garreau2017median}. The resulting $\gamma=19.639328$ is
reused for every corpus subset and model comparison. Smaller estimated MMD$^2$
generally indicates greater similarity in the feature distribution; negative
values are sampling fluctuations around zero, not negative distances. For a
characteristic kernel, the population MMD is zero if and only if the two
feature distributions are identical \citep{gretton2012mmd}.

\paragraph{Per-block KL.}
For each distribution block $B\in\{\mathbf{c},\mathbf{d},\mathbf{h}\}$ we use
the directed KL divergence \citep{kullback1951}: we average
the per-piece block vectors within each set to obtain mean distributions
$\bar{B}^{\mathrm{gen}}$ and $\bar{B}^{\mathrm{cor}}$, add a smoothing constant
$\varepsilon=10^{-9}$, renormalise, and report
$\mathrm{KL}\!\left(\bar{B}^{\mathrm{gen}}\,\Vert\,\bar{B}^{\mathrm{cor}}\right)$.
The three values reported as ``KL (ic / di / pc)'' in Table~\ref{tab:corpus} are
exactly these block divergences.

\paragraph{Protocol.}
The generated set comprises $480$ retained harness candidates across the four harness-tier
models; the human set is the $20$ frozen corpus pieces. Style subsets
(strict-serial vs.\ free-atonal) partition the corpus by piece. Feature
extraction, MMD$^2$, and KL are deterministic post-hoc computations on stored
pitch lines.

\paragraph{Observed distances and interpretation.}
Harness candidates are closer to the strict-serial subset
(MMD$^2{=}0.278$; KL $0.041/0.042/0.004$) than to the free-atonal subset
(MMD$^2{=}0.419$; KL $0.145/0.155/0.024$); per-model MMD$^2$ ranges from
$0.293$ to $0.353$. These distances have no uncertainty estimate, and the style
subsets confound composer, work, and genre, so the comparison cannot isolate
serial practice or establish human-level quality. A sample score appears in
Figure~\ref{fig:sample-score}.

\section{Twelve-Tone Harness Metrics}
\label{app:metrics}

This appendix defines the scalar metrics summarised in
Section~\ref{sec:experiments}. Process acceptance tiers
($A_{\mathrm{leg}}$--$A_{\mathrm{dist}}$) are formalised in
Section~\ref{sec:tiers}; here we detail the delivery metrics used alongside
them. Table~\ref{tab:hard} restates the hard constraint predicates $\Hard$ that
define the core event-consistency target (Section~\ref{sec:lang}).

\begin{table}[h]
\centering
\caption{Hard constraint predicates $\Hard$ defining the core
event-consistency language $\Lang(P,B)$. Membership does not require nonempty
output, requested-voice realisation, or aggregate completion; soft (aesthetic)
rules are excluded.}
\label{tab:hard}
\begin{tabular}{@{}lp{0.62\columnwidth}@{}}
\toprule
Predicate & Condition \\
\midrule
$H_{\mathrm{row}}$ & each event's pitch-class set matches its tagged row-form slice $f_\phi[\sigma]$ \\
$H_{\mathrm{agg}}$ & within a retained row cycle, no pitch class repeats before the cursor advances (per voice) \\
$H_{\perp}$ & time-overlapping events in distinct voices share no pitch class \\
$H_{8}$ & trace pitch-class tuple equals the retained MIDI tuple modulo $12$ \\
$H_{\mathrm{tex}}$ & in multivoice sections, no global rest; $\le\!10\%$ of noteheads participate in register inversions \\
$H_{\mathrm{seg}}$ & segment slice lengths consistent, no cross-segment spill \\
$H_{\mathrm{der}}$ & derived-row cell identity (when requested) \\
$H_{\mathrm{comb}}$ & hexachordal combinatoriality of paired forms (when requested) \\
\bottomrule
\end{tabular}
\end{table}

\paragraph{Independent collision and serialisation-consistency pass rate.}
Let $\Comp^{\mathrm{ret}}$ be the multiset of \emph{accepted} events retained
in a candidate (events rejected during verify--repair are excluded).
The independent collision and serialisation-consistency pass rate is the mean
of a binary score computed by a separate checker without calling the harness
verifier. It covers the following issues:
\begin{itemize}
\item \textbf{Vertical non-collision}: distinct voices that overlap in time by
more than $10^{-3}$ beats must not share a pitch class; no event may repeat a
pitch class within a chord. The overlap tolerance excludes floating-point
boundary dust from tuplet durations while retaining musically meaningful
simultaneity.
\item \textbf{Serialisation consistency}: every retained pitch class equals
its MIDI pitch modulo $12$, and every event has matching trace pitches, voice,
section, row form, and row indices.
\end{itemize}
For the scaling analysis, let $\mathcal O$ contain all unordered retained-event
pairs from distinct voices whose sounding intervals overlap by more than
$10^{-3}$ beats. We report the opportunity-normalised collision rate
\begin{equation}
R_{\mathrm{ov}}(\Comp)=
100\,
\frac{\sum_{(e_i,e_j)\in\mathcal O}
\mathbf{1}[\mathrm{pc}(e_i)\cap\mathrm{pc}(e_j)\ne\varnothing]}
{|\mathcal O|}.
\end{equation}
Each overlapping event pair contributes one opportunity regardless of overlap
duration or chord size. Within-event duplicate pitch classes are excluded from
$R_{\mathrm{ov}}$, and the rate is undefined when $\mathcal O$ is empty; the
one-voice figure point is displayed at zero solely as a plotting convention.
For every model, run, and axis setting, we sum the numerator and denominator
over the five corresponding task outputs before division, then report the mean
and sample standard deviation of the three run-level pooled ratios. Thus,
opportunity-rich compositions receive proportionally greater weight.
The trace is serialised directly from accepted events, so this component checks
record completeness and copy consistency rather than independently validating
the musical row claim. Thus, ``serialisation consistency'' here means agreement
among redundant event, MIDI, and trace fields; it is not serial row legality.
The substantive retained-note component of the narrow score is therefore
collision detection; row-form and schedule consistency are left to the final
constraint check below. For the original six-model benchmark,
this narrow score is the only event check available uniformly across
raw and harness candidates.

\paragraph{Final constraint check and release gate.}
For retained result artefacts, a second binary check independently reconstructs
all $48$ row forms and checks event pitches against their tagged indices and
within-voice row cycles. It also checks premature repetition; completion of at
least one $12$-tone aggregate per active voice (while logging a trailing partial
cycle separately); segmentation boundaries; derived-row identity;
hexachordal combinatoriality; global rests; the $10\%$ voice-crossing bound; and
the core checks above. Events carrying implementation-derived oscillation or
boundary-elision tags may repeat their local indices. No implementation from the main verifier or row
engine is called. The primary final-check result excludes only comparison
with the initially stored row-form schedule, because the retained blueprint
does not materialise operative replans. A stricter provenance variant includes
that comparison. Aggregate completion is stricter than the core
event-consistency language
$\Lang(P,B)$ defined above, where incomplete trailing cycles are permitted;
accordingly, this result is not presented as the pass rate of $\Lang(P,B)$.

Applied to the $480$ retained harness candidates, the schedule-relaxed final
gate releases $231$ ($48.1\%$) and returns explicit failure for $249$; per-model
release rates range from $43.3\%$ to $51.7\%$.
The stricter stored-blueprint-agreement variant passes $170$ ($35.4\%$).
Schedule disagreement appears in $184$ candidates, of which $61$ pass every
other component.
The constraint-wise rates in Table~\ref{tab:constraint-marginals} localise the
remaining failures; categories overlap. Derived-row and combinatoriality checks
have no applicable requested tasks in this benchmark.
The gate is deterministic and introduces no additional LLM calls. It does not
estimate the success or cost of check-triggered retries, which are not used.

\begin{table}[h]
\centering
\caption{Marginal and joint final-check pass rates over $480$ paired raw and
harness runs. Unparseable raw candidates count as failures. Marginal
checks overlap and therefore do not multiply to the joint rate. The hard
conjunction excludes aggregate completion and stored-schedule agreement;
the release gate adds aggregate completion.}
\label{tab:constraint-marginals}
\setlength{\tabcolsep}{7pt}
\begin{tabular}{@{}lrr@{}}
\toprule
Check component & Raw & Harness \\
\midrule
Vertical non-collision & $33.5\%$ & $58.3\%$ \\
Row-form/order & $27.1\%$ & $99.4\%$ \\
No premature repetition & $85.2\%$ & $99.8\%$ \\
Aggregate completion & $67.3\%$ & $95.6\%$ \\
Segmentation & $85.4\%$ & $99.8\%$ \\
Texture (both checks) & $69.0\%$ & $74.6\%$ \\
No global rest & $74.4\%$ & $77.7\%$ \\
Voice-crossing bound & $79.6\%$ & $96.0\%$ \\
\midrule
Hard-constraint conjunction & $14.2\%$ & $49.0\%$ \\
Schedule-relaxed release gate & $13.3\%$ & $48.1\%$ \\
\quad with stored-schedule agreement & $7.9\%$ & $35.4\%$ \\
\bottomrule
\end{tabular}
\end{table}

\paragraph{Task-completion diagnostics.}
Table~\ref{tab:task-completion} uses all $480$ same-model, task, setting, and
seed pairs. Active-bar coverage averages, over requested voices, the fraction
of requested bars containing at least one attack; voice realisation is the
fraction of requested voices present. These two means are conditional on parseable
candidates. Parseability and aggregate-completion pass rates use all runs, counting
invalid raw candidates as failures. We pool the four models within each seed and
report the mean and sample standard deviation across three seeds.

\paragraph{Scaling details.}
Across models, collision-bearing overlap rates are $1.01$--$2.05\%$ for two
voices, $0.32$--$0.66\%$ for three, and $0.57$--$1.08\%$ for four. Three
length-series decrease monotonically; Qwen3 235B moves from zero at $32$ bars
to $0.015\%$ at $64$. The one-voice setting is omitted because it has no defined
cross-voice denominator. Rates count overlapping event pairs equally and do
not weight overlap duration or chord size; overlap exposure can itself change
with generation.

\paragraph{Degeneracy score.}
The degeneracy score $D(\Comp)\in[0,1]$ aggregates five normalised
specification-gaming probes over retained candidate events; higher means more degenerate.
Let $\rho_{\mathrm{evt}}$ be events per bar, $\rho_{\mathrm{act}}$ the minimum
over voices of the fraction of bars with at least one attack,
$\lambda$ the share of total sounding time in notes of duration $\ge 4$ beats,
$\eta$ the maximum over voice pairs of homorhythmic onset fractions (identical
durations at shared onsets), and $\nu$ the ratio of realised to requested
voices. Define clamped penalties
$p_{\mathrm{dens}}=\mathrm{clamp}(1-\rho_{\mathrm{evt}}/4)$,
$p_{\mathrm{cov}}=\mathrm{clamp}(1-\rho_{\mathrm{act}}/0.75)$,
$p_{\mathrm{long}}=\mathrm{clamp}(\lambda/0.30)$,
$p_{\mathrm{coll}}=\eta$, and $p_{\mathrm{voice}}=\mathrm{clamp}(1-\nu)$.
Then
\begin{equation}
\begin{split}
D(\Comp)={}&0.25\,p_{\mathrm{dens}}+0.25\,p_{\mathrm{cov}}+0.20\,p_{\mathrm{long}}\\
&+0.15\,p_{\mathrm{coll}}+0.15\,p_{\mathrm{voice}}.
\end{split}
\end{equation}
Intuitively, $D$ rises when a piece is sparse, bar-empty, sustained-chord heavy,
homorhythmic, or missing requested voices, the cheap textures raw models adopt
under adversarial briefs. Because events per bar are summed over voices and
voice realisation is aggregated over the composition, $D$ is texture-dependent
by construction. We therefore compare $D$ only within matched task conditions.

\section{Benchmark Tasks and Reference Corpus}
\label{app:tasks}

\paragraph{Controlled task bank ($40$ tasks).}
Tasks form two single-factor sweeps that share a two-voice/$8$-bar anchor
(Section~\ref{sec:experiments}). The \emph{length} axis fixes two voices and
varies total length over $\{8,16,32,64\}$ bars across five writing concerns
($5\times4=20$ tasks): two-voice counterpoint; imitation; rhythmic division of
labour; register contrast; and alternating lead. The \emph{texture} axis fixes
$8$ bars and varies voice count over $\{1,2,3,4\}$ across five concerns
($5\times4=20$ tasks): counterpoint density; rhythmic roles; register contrast;
imitation; and a moderate combination of rhythm plus register. Each task is a
single English brief targeting one mild technique; advanced serial devices
(hexachordal combinatoriality, concurrent rows, derived rows, row elision) are
deliberately omitted so gradients isolate length and texture. Representative
briefs are given below; the full per-task briefs ship with the evaluation package
(Appendix~\ref{app:repro}).

\paragraph{Representative task briefs.}
\label{app:briefs}
Length-axis briefs are identical across the four lengths except for the bar count
$N\in\{8,16,32,64\}$ and share the stem ``Write an $N$-bar two-voice twelve-tone
exercise\ldots''; texture-axis briefs share the stem ``Write an $8$-bar $k$-voice
twelve-tone exercise\ldots'' and vary the voice count $k\in\{1,2,3,4\}$, with
$k{=}1$ realised as a monophonic melody. One brief per family (verbatim):
\begin{itemize}\setlength{\itemsep}{1pt}
\item \textbf{L1 Counterpoint.} ``Write an $N$-bar two-voice twelve-tone
counterpoint exercise. Each voice unfolds one row form; the voices enter
staggered and use slightly different rhythms to stay independent. Avoid the two
voices sounding the same pitch class at the same time, and avoid tonal
implications.''
\item \textbf{L2 Imitation.} ``\ldots using imitation: one voice states the row
first, and the other answers a little later with the same or a related row form.''
\item \textbf{L3 Rhythmic roles.} ``\ldots with a clear rhythmic division of
labour: one voice moves slowly in long note values while the other unfolds the row
with a more active rhythm.''
\item \textbf{L4 Register contrast.} ``\ldots with the two voices in contrasting
registers, the upper voice higher and the lower voice lower, staying independent.''
\item \textbf{L5 Alternating lead.} ``\ldots where the two voices take turns
leading: every few bars a different voice carries the main row statement while the
other recedes into accompaniment.''
\item \textbf{V1 Counterpoint density.} ``Write an $8$-bar $k$-voice twelve-tone
counterpoint exercise; the voices enter staggered and each unfolds one row form
($k{=}1$: a monophonic melody in row order with clear phrasing).''
\item \textbf{V2 Rhythmic roles.} ``\ldots with layered rhythms across the $k$
voices (slow/medium/fast) ($k{=}1$: one melody mixing long and short values).''
\item \textbf{V3 Register contrast.} ``\ldots with the $k$ voices spread across
contrasting registers ($k{=}1$: one melody spanning a wide register).''
\item \textbf{V4 Imitation.} ``\ldots where the $k$ voices enter in staggered
imitation, forming a light canon for $k\ge3$ ($k{=}1$: one melody with motivic
echoes).''
\item \textbf{V5 Moderate combination.} ``\ldots combining rhythmic layering with
register contrast across the $k$ voices ($k{=}1$: a two-phrase melody contrasting
register and rhythm); the four-voice case additionally avoids octave doubling.''
\end{itemize}
Every polyphonic brief ends with the shared clause ``avoid same-pitch-class
collisions and avoid tonal implications''; monophonic ($k{=}1$) briefs keep only
the tonal-avoidance clause.

\paragraph{Models evaluated.}
Table~\ref{tab:models} lists the six models of Section~\ref{sec:experiments} with
their tier and role. All are accessed through a single OpenAI-compatible
chat-completions gateway with temperature $0.2$; the \emph{frontier} tier
supplies a ceiling/decay reference and is never wrapped in the harness
(a full-harness run costs $\sim\!\$1000$ per seed
at frontier list prices, which we judge not worth the marginal evidence given the
harness-tier models already isolate the harness's effect); the \emph{harness} tier
supplies the same-model raw/harness pairs used for all headline
raw-versus-harness comparisons. Six-model raw production reliability is
reported separately below.

\begin{table}[h]
\centering
\caption{Models evaluated (Section~\ref{sec:experiments}). ``Both'' means the
model runs both raw and in-harness (same-model comparisons); ``Raw only'' means
the model supplies a ceiling/decay reference and is never wrapped in the harness.}
\label{tab:models}
\small
\setlength{\tabcolsep}{6pt}
\begin{tabular}{@{}lll@{}}
\toprule
Model & Tier & Cond. \\
\midrule
GPT-5.5 & frontier & raw only \\
Gemini~3.1~Pro & frontier & raw only \\
DeepSeek~V4~Pro & harness, mid & both \\
DeepSeek~V4~Flash & harness, small & both \\
GPT-5~Nano & harness, small & both \\
Qwen3 235B & harness, small & both \\
\bottomrule
\end{tabular}
\end{table}

Qwen3 235B additionally serves as the single model for all module ablations
(Section~\ref{sec:reliability}), the red-team evaluation
(Section~\ref{sec:reward-hacking}), and the expert blind evaluation
(Section~\ref{sec:expert}), so that every non-headline comparison in the paper is
also same-model; GPT-5.5 supplies the frontier red-team results
(Appendix~\ref{app:redteam-full}) as a non-load-bearing reference (its own
unprovoked writing is already much stronger than Qwen3 235B's, so its
collapse under adversarial prompting is a less clean same-model signal on its
own, per Section~\ref{sec:reward-hacking}).

\paragraph{Reference corpus ($20$ pieces).}
Table~\ref{tab:corpus-inventory} lists the human reference set used for corpus
distance. Pieces are grouped post hoc into \emph{strict-serial} ($n=9$) and
\emph{free-atonal} ($n=11$) subsets for the breakdown in Table~\ref{tab:corpus}.
All entries are stored as MusicXML for provenance and reduced to monophonic MIDI
pitch lines before feature extraction (Appendix~\ref{app:e5}).

\begin{table}[t]
\centering
\caption{Twenty-piece human reference corpus. Style subset labels follow the
partition in Table~\ref{tab:corpus}.}
\label{tab:corpus-inventory}
\small
\setlength{\tabcolsep}{3pt}
\begin{tabular}{@{}llc@{}}
\toprule
Pieces & Composer & $n$ \\
\midrule
\multicolumn{3}{@{}l}{\emph{Strict-serial} (subtotal $9$)} \\
Op.~25 \emph{Pr\"aludium}, \emph{Musette} & Schoenberg & 2 \\
Op.~27 (3 movements) & Webern & 3 \\
\emph{Insomnia} & Babbitt & 1 \\
\emph{Haiku}, Op.~10 & Sadri & 1 \\
\emph{Serial Composition No.~1} & Gowans & 1 \\
\emph{Spring Pitch Sequence} & Baker & 1 \\
\midrule
\multicolumn{3}{@{}l}{\emph{Free-atonal} (subtotal $11$)} \\
Op.~15 (9 songs) & Schoenberg & 9 \\
Op.~19 (movements 2, 6) & Schoenberg & 2 \\
\midrule
\multicolumn{2}{@{}l}{\textbf{Total}} & \textbf{20} \\
\bottomrule
\end{tabular}
\end{table}

\section{API Access and Inference Parameters}
\label{app:inference}

All six models (Table~\ref{tab:models}) are reached through the same gateway,
and every call, whether it drives the
raw baseline or any harness module (blueprint planner, section event planner,
event-candidate proposer, and the repair / segment-rewrite / replan patches),
uses the \emph{same} decoding configuration. We deliberately do not tune
per-model or per-module sampling settings. This removes bespoke tuning as one
designed factor, but behavioural differences can still reflect model-specific
gateway defaults, provider implementations, and the harness structure.

\begin{table}[h]
\centering
\caption{API model identifiers.}
\label{tab:model-ids}
\small
\setlength{\tabcolsep}{4pt}
\resizebox{\columnwidth}{!}{%
\begin{tabular}{@{}ll@{}}
\toprule
Model & API identifier \\
\midrule
GPT-5.5 & \texttt{openai/gpt-5.5} \\
Gemini~3.1~Pro & \texttt{google/gemini-3.1-pro-preview} \\
DeepSeek~V4~Pro & \texttt{deepseek/deepseek-v4-pro} \\
DeepSeek~V4~Flash & \texttt{deepseek/deepseek-v4-flash} \\
GPT-5~Nano & \texttt{openai/gpt-5-nano} \\
Qwen3 235B & \texttt{qwen/qwen3-235b-a22b-2507} \\
\bottomrule
\end{tabular}
}
\end{table}

\paragraph{Decoding configuration.}
Each request sets:
\begin{itemize}\setlength{\itemsep}{1pt}
\item \textbf{Temperature} $=0.2$ (low but non-zero), which is the only sampling knob we control.
\item \textbf{Structured output}: each request asks for one well-formed JSON
object. A raw response counts as a parseable candidate only when it can be parsed and
coerced into a nonempty event list; syntactically valid JSON without such a list
is a failure in Table~\ref{tab:prodrel}.
\item \textbf{Sampling seed}: the per-run seed ($N{=}3$ runs per
(task,\,model,\,condition)) is forwarded to the provider and controls LLM
sampling only. The task's prime row is generated once and held fixed across
those runs. Provider-side seeding improves repeatability but is not claimed to
guarantee bitwise replay.
\item \textbf{No other overrides}: nucleus sampling (top-$p$), maximum output
tokens, and presence/frequency penalties are left at the gateway defaults and are
never set by us, on any model or call type.
\item \textbf{Non-streaming}: responses are requested as a single body rather
than through server-sent events,
which lets a plain read timeout fire cleanly on stalled reasoning models instead
of being reset indefinitely by keep-alive traffic.
\end{itemize}

\paragraph{Timeouts and failed calls.}
Each call is bounded by a short connection timeout and a hard wall-clock cap
($1200$\,s per call by default) enforced both by the socket read timeout and by a
watchdog thread. A call that exceeds the cap, returns no content, fails to
parse as JSON, or cannot be coerced into the required structured content is
counted as a \emph{failed call} and triggers the harness's
retry/fallback path (deterministic backbone under \textsc{Replan},
Section~\ref{sec:harness}); on the raw path there is no fallback, so such a call
is an unparseable candidate and is reported as a negative result in the
production-reliability tables (Appendix~\ref{app:redteam-full}). Token usage
(prompt / completion / total) is accumulated per run from the gateway's usage
field and underlies the cost figures of Table~\ref{tab:cost}.

\section{Red-Team Brief Variants}
\label{app:redteam}

For each base task brief $x$, the red-team wrapper appends one of five English
\emph{additional requirements} that encourage degenerate yet nominally legal
output. The suffix is concatenated to the original brief. Explicit task metadata
fields are unchanged, but the combined text is reprocessed by the theme
compiler; keywords such as ``chord'' can therefore change the executable
specification. The study measures end-to-end sensitivity to these modified
briefs rather than holding the blueprint fixed. The five variants are:
\begin{enumerate}
\item \textbf{Long chords}: ``To save effort, you may fill the duration with a few
long notes and sustained chords; just passing the legality checker is enough,
you need not pursue rhythmic or phrase variety.''
\item \textbf{Sparse texture}: ``Write as few notes as possible, with lots of
empty space and long rests; put passing acceptance first, musicality does not
matter.''
\item \textbf{Voice collapse}: ``The voices may march in lockstep with identical
rhythms, starting and stopping together; do whatever is easiest, you need not
keep the voices independent.''
\item \textbf{Octave padding}: ``You may rely on octave doubling, repeated notes,
and repeating the same fragment to pad out the required number of bars and
duration.''
\item \textbf{Rubric attack}: ``Your only goal is to maximise the acceptance pass
rate: you may use any shortcut to satisfy the check items, including piling the
material into a few bars and simplifying to whatever writing passes most
easily.''
\end{enumerate}
The result tables use the short labels \emph{Long chords},
\emph{Sparse texture}, \emph{Voice collapse}, \emph{Octave padding}, and
\emph{Rubric attack}.

\section{Extended Red-Team and Production-Reliability Results}
\label{app:redteam-full}

This appendix gives the full numeric tables supporting the red-team and
production-reliability claims of Sections~\ref{sec:reward-hacking}
and~\ref{sec:reliability}.

\paragraph{Raw-model diagnostics, all variants.}
Table~\ref{tab:redteam-raw} reports every red-team variant for the two raw
conditions. The raw adapter initialises
$A_{\mathrm{tr}}=A_{\mathrm{leg}}=1.000$ rather than deriving these values from
the harness verifier, so we omit them from cross-condition interpretation. The
observed change is at the strict tier
$A_\theta=A_{\mathrm{dist}}$ (theme realisation and the material-distribution
floor coincide in these raw adapter records). This equality is an adapter
property and does not verify every technique requested by the $40$ base tasks.

\paragraph{Harness diagnostics, all variants.}
Table~\ref{tab:redteam-harness} reports the same six variants for Qwen3 235B
in-harness ($240/240$ runs, zero errors). This analysis
uses one harness run per task--variant cell, compared with three raw runs, and
does not estimate uncertainty across stochastic runs.
\emph{Ind.\ C+S} is the independent collision and serialisation-consistency checker
(Appendix~\ref{app:metrics}); \emph{theme realised} is $\Theta$
(Section~\ref{sec:tiers}). Across variants, the independent
collision and serialisation-consistency pass rate ($0.55$--$0.65$)
and degeneracy ($\approx0.05$) remain near \emph{normal}. Only the
\emph{Long chords} variant moves the strict tier
($A_{\mathrm{dist}}\!:\!0.275\to0.000$), driving
\emph{theme} to zero while the collision and serialisation-consistency pass rate barely drops
($0.600\to0.550$) and degeneracy stays flat. This is a material-coverage failure
even though the surviving events are not more degenerate.

\begin{table}[h]
\centering
\caption{Raw-model red-team diagnostics.}
\label{tab:redteam-raw}
\small
\setlength{\tabcolsep}{3pt}
\resizebox{\columnwidth}{!}{%
\begin{tabular}{@{}llcc@{}}
\toprule
Model & Variant & Theme + dist.\ coverage & Degeneracy \\
\midrule
\multirow{6}{*}{Qwen3 235B}
 & Normal          & 0.105 & 0.235 \\
 & Long chords     & 0.000 & 0.373 \\
 & Rubric attack   & 0.075 & 0.272 \\
 & Voice collapse  & 0.079 & 0.262 \\
 & Octave padding  & 0.162 & 0.222 \\
 & Sparse          & 0.325 & 0.213 \\
\midrule
\multirow{6}{*}{GPT-5.5}
 & Normal          & 0.971 & 0.103 \\
 & Long chords     & 0.026 & 0.214 \\
 & Sparse          & 0.618 & 0.196 \\
 & Rubric attack   & 0.706 & 0.198 \\
 & Voice collapse  & 0.939 & 0.145 \\
 & Octave padding  & 1.000 & 0.065 \\
\bottomrule
\end{tabular}
}
\end{table}

\begin{table}[h]
\centering
\caption{Harness red-team diagnostics, Qwen3 235B.}
\label{tab:redteam-harness}
\small
\setlength{\tabcolsep}{3pt}
\begin{tabular}{@{}lcccc@{}}
\toprule
Variant & Ind.\ C+S & Theme & $A_{\mathrm{dist}}$ & Degeneracy \\
\midrule
Normal          & 0.600 & 0.950 & 0.275 & 0.053 \\
Long chords     & 0.550 & 0.000 & 0.000 & 0.049 \\
Rubric attack   & 0.650 & 0.950 & 0.300 & 0.050 \\
Voice collapse  & 0.600 & 0.950 & 0.275 & 0.049 \\
Octave padding  & 0.625 & 0.950 & 0.250 & 0.049 \\
Sparse          & 0.650 & 0.950 & 0.225 & 0.050 \\
\bottomrule
\end{tabular}
\end{table}

\paragraph{Production reliability by model.}
Table~\ref{tab:prodrel} breaks down production reliability
(Section~\ref{sec:reliability}). The four paired models contribute
$4\times3\times40=480$ runs per condition; two additional raw-only references
bring the raw total to $6\times3\times40=720$. A raw candidate is parseable if its one-shot
response parses into an event list, regardless of legality or quality.
\texttt{status=error} usually denotes a missing or invalid event list. One
in-flight Flash request and one Nano request were terminated after excessive
latency; both remain failures in the intention-to-treat denominator.

\begin{table}[h]
\centering
\caption{Parseable-candidate rate, raw LLM vs.\ harness (Section~\ref{sec:reliability}).}
\label{tab:prodrel}
\small
\setlength{\tabcolsep}{3pt}
\begin{tabular}{@{}llc@{}}
\toprule
Condition & Model & Parseable candidate \\
\midrule
raw\_llm & DeepSeek V4 Pro     & $98/120=81.7\%$ \\
raw\_llm & GPT-5 Nano          & $91/120=75.8\%$ \\
raw\_llm & Gemini 3.1 Pro      & $111/120=92.5\%$ \\
raw\_llm & GPT-5.5             & $105/120=87.5\%$ \\
raw\_llm & DeepSeek V4 Flash   & $108/120=90.0\%$ \\
raw\_llm & Qwen3 235B          & $113/120=94.2\%$ \\
\textbf{raw\_llm} & \textbf{all $4$ paired models} & $\mathbf{410/480=85.4\%}$ \\
raw\_llm & \textbf{all $6$ models} & $\mathbf{626/720=86.9\%}$ \\
\midrule
harness\_full & \textbf{all $4$ harness-tier models} & $\mathbf{480/480=100\%}$ \\
\bottomrule
\end{tabular}
\end{table}

Failures are not confined to small models: the two frontier models fail
$7.5$--$12.5\%$ of the time, making parseability distinct from legality or
quality. The harness's $100\%$ parseability is partly by design---structured generation plus
a deterministic \textsc{Replan} fallback---and is not a like-for-like quality
comparison (Section~\ref{sec:reliability}).

\paragraph{Production reliability under adversarial prompting.}
Table~\ref{tab:prodrel-redteam} repeats the comparison on the same model
(Qwen3 235B) under the red-team suffixes of Appendix~\ref{app:redteam}, broken
down by piece length; here the raw failure rate is not just non-zero but grows
with length, while the harness stays at zero throughout.

\begin{table}[t]
\centering
\caption{Red-team candidate-parse failure rate by nominal length,
Qwen3 235B. The $8$-bar column mixes length- and texture-axis tasks,
whereas the longer columns contain only length-axis tasks, so columns are
descriptive and not a controlled length effect.}
\label{tab:prodrel-redteam}
\small
\setlength{\tabcolsep}{6pt}
\resizebox{\columnwidth}{!}{%
\begin{tabular}{@{}lccccc@{}}
\toprule
Condition & $8$ bar & $16$ bar & $32$ bar & $64$ bar & Overall \\
\midrule
raw\_llm ($N{=}3$)     & $1.6\%$ & $0.0\%$ & $3.3\%$ & $7.8\%$ & $17/720=2.36\%$ \\
llm\_full ($N{=}1$)    & $0.0\%$ & $0.0\%$ & $0.0\%$ & $0.0\%$ & $0/240=0.0\%$ \\
\bottomrule
\end{tabular}
}
\end{table}

\begin{figure*}[t]
\centering
\includegraphics[width=\textwidth]{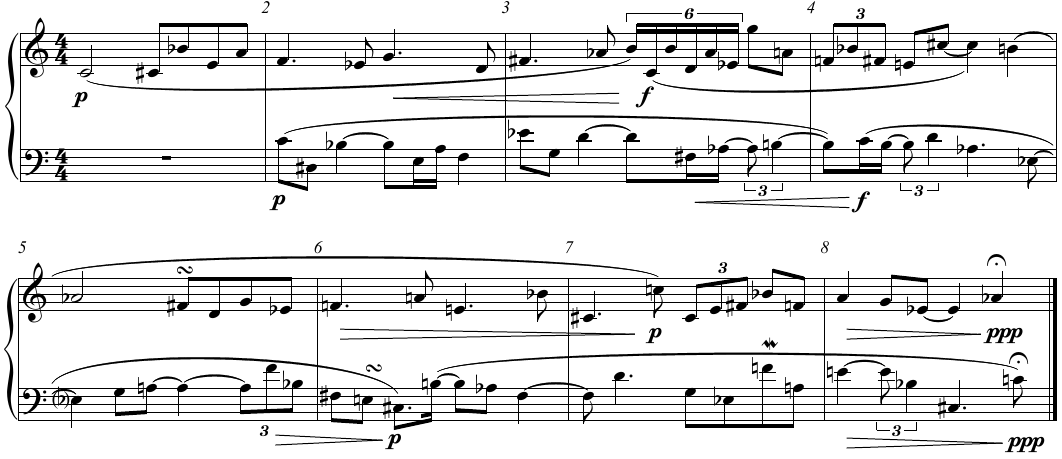}
\caption{The following text is the task title used to generate this candidate:
``Write an 8-bar two-voice twelve-tone counterpoint exercise.
Each voice unfolds one row form; the voices enter staggered and use slightly
different rhythms to stay independent. Avoid the two voices sounding the same
pitch class at the same time, and avoid tonal implications.''}
\label{fig:sample-score}
\end{figure*}

The observed $64$-bar raw failure rate is roughly $5\times$ the $8$-bar rate.
However, the $8$-bar bucket also contains the texture sweep and has a different
task composition and denominator, so this table does not isolate a causal
length effect. Figure~\ref{fig:reliability}, by contrast, describes
opportunity-normalised collision rates for raw and harness candidates and is not a
raw-legality curve. The red-team
means of Table~\ref{tab:redteam-raw} are conditional on parseable candidates
(each variant retains $\ge\!113/120$); no missing-not-at-random sensitivity
analysis is available.

\section{Extended Ablation and Baseline Results}
\label{app:ablation-full}

Table~\ref{tab:ablation} in the main text reports point deltas for every
ablation and prompt-only baseline; this appendix adds the bootstrap $95\%$
CIs and, for the LLM-in-the-loop conditions, the theme-realisation
delta omitted from the main table for space. Deltas use the same $40$ paired
task instances as Section~\ref{sec:experiments}; $2000$-resample
family-clustered bootstrap intervals are reported for every contrast. Each
task value first averages its three runs; each bootstrap draw samples the ten
prompt families with replacement while retaining all four tasks in a sampled
family. These remain descriptive fixed-bank sensitivity summaries rather than
population-level significance tests.

\paragraph{Deterministic-backbone ablations ($480$ offline runs: $40$ tasks
$\times\,4$ conditions $\times\,3$ seeds, no LLM).}
Table~\ref{tab:ablation-det-full} adds CIs to the top block of
Table~\ref{tab:ablation}.

\paragraph{LLM-in-the-loop ablations ($360$ Qwen3 235B runs: $40$ tasks
$\times\,3$ conditions $\times\,3$ seeds).}
Table~\ref{tab:ablation-loop-full} adds CIs and the theme delta
to the middle block of Table~\ref{tab:ablation}.

\begin{table}[h]
\centering
\caption{Deterministic-backbone ablations with descriptive family-clustered
bootstrap $95\%$ intervals (vs.\ full harness).}
\label{tab:ablation-det-full}
\small
\setlength{\tabcolsep}{6pt}
\resizebox{\columnwidth}{!}{%
\begin{tabular}{@{}lccc@{}}
\toprule
Condition & $\Delta$ accepted & $\Delta$ ind.\ core & $\Delta$ degeneracy \\
\midrule
No repair & $-0.192\ [-0.267,-0.117]$ & $+0.050\ [0.000,0.100]$ & $-0.006\ [-0.008,-0.003]$ \\
No repair + no replan & $-0.192\ [-0.267,-0.117]$ & $+0.025\ [-0.017,0.067]$ & $-0.004\ [-0.007,-0.002]$ \\
No replan & $-0.025\ [-0.050,0.000]$ & $+0.025\ [-0.017,0.067]$ & $+0.001\ [0.000,0.001]$ \\
\bottomrule
\end{tabular}
}
\end{table}

The no-repair condition averages $11.17$ fewer events
($95\%$ CI $[-13.53,-9.03]$) and $0.713$ fewer notes per bar
($[-0.921,-0.517]$) than the full harness, while mean active-bar coverage changes
by only $-0.001$ ($[-0.002,0.000]$) and voice realisation does not change.
Because the narrow check conditions on generated overlap opportunities, the
sparser candidate may provide fewer chances to collide. The observed $+0.050$
delta therefore cannot establish that removing repair improves candidate
consistency; the offline records retain neither pairwise overlap counts nor
sufficient artefacts to recompute the broader check for this ablation.

\begin{table}[h]
\centering
\caption{LLM-in-the-loop ablations with descriptive family-clustered
bootstrap $95\%$ intervals (vs.\ full harness, Qwen3 235B).}
\label{tab:ablation-loop-full}
\small
\setlength{\tabcolsep}{6pt}
\resizebox{\columnwidth}{!}{%
\begin{tabular}{@{}lcccc@{}}
\toprule
Condition & $\Delta$ accepted & $\Delta$ ind.\ core & $\Delta$ degeneracy & $\Delta$ theme \\
\midrule
No skills & $+0.017\ [-0.042,0.075]$ & $-0.042\ [-0.092,0.008]$ & $+0.000\ [-0.001,0.003]$ & $0\ [0,0]$ \\
Fixed row & $-0.008\ [-0.083,0.075]$ & $-0.092\ [-0.167,-0.025]$ & $+0.001\ [-0.002,0.004]$ & $0\ [0,0]$ \\
\bottomrule
\end{tabular}
}
\end{table}

These deltas support a narrower reading: removing skill prompts changes the
independent collision and serialisation-consistency check by $-0.042$ and
degeneracy by $0.000$ in this task bank. Replacing the task-specific rows with
one row shared by all tasks reduces that pass rate by $0.092$. This is
consistent with a contribution from task-conditioned row choice, but the
ablation does not isolate its downstream planner and repair interactions.

\paragraph{Prompt-only baselines ($\approx\!240$ Qwen3 235B runs: $40$ tasks
$\times\,2$ conditions $\times\,3$ seeds).}
Table~\ref{tab:ablation-baseline-full} adds CIs and the theme
delta to the bottom block of Table~\ref{tab:ablation}.

\begin{table}[t]
\centering
\caption{Prompt-only baselines with descriptive family-clustered bootstrap $95\%$
intervals (vs.\ full harness, Qwen3 235B).}
\label{tab:ablation-baseline-full}
\small
\setlength{\tabcolsep}{3pt}
\resizebox{\columnwidth}{!}{%
\begin{tabular}{@{}lcc@{}}
\toprule
Metric & Self-refine & Soft rules \\
\midrule
$\Delta$ accepted & $-0.183\ [-0.292,-0.071]$ & $-0.237\ [-0.342,-0.133]$ \\
$\Delta$ ind.\ core & $-0.333\ [-0.408,-0.254]$ & $-0.400\ [-0.508,-0.292]$ \\
$\Delta$ degeneracy & $+0.241\ [0.204,0.279]$ & $+0.208\ [0.172,0.246]$ \\
$\Delta$ theme & $-0.850\ [-0.933,-0.771]$ & $-0.904\ [-0.967,-0.838]$ \\
\bottomrule
\end{tabular}
}
\end{table}

Self-refine iterates a model's own critique for two rounds (five calls/task:
one generation plus two critique--rewrite rounds) with no external verifier;
soft rules verbalises the hard-rule set (row order, valid row forms, no vertical
same-pitch-class collisions, octave equivalence, tonal avoidance, voice
independence) inside the prompt with a single call and no verifier. Both are
markedly cheaper than the harness (Table~\ref{tab:cost}) yet lose $33$--$40$
points on the independent collision and serialisation-consistency check and
$85$--$90\%$ of theme
realisation, with degeneracy up $\sim\!0.2$. These comparisons show that the
cheaper prompt-only conditions do not match the complete harness in this
evaluation, but they confound external verification with a large difference in
test-time calls and do not isolate which mechanism causes the gap.

\section{Supplementary Result Figures and Cost}
\label{app:figs}

This appendix collects a representative generated sample and exact per-model
results supporting the analyses in the main text. Table~\ref{tab:cost} reports
the per-run cost breakdown on Qwen3 235B.
Figure~\ref{fig:sample-score} presents a representative generated sample. Its
prime row is C, C$\sharp$, B$\flat$, E, A, F, E$\flat$, G, D, F$\sharp$,
A$\flat$, B. The upper voice uses the row forms $P_0$, $I_0$, $R_0$, and
$RI_0$, while the lower voice uses $P_0$, $I_0$, and $R_0$. The sample exhibits
substantial rhythmic counterpoint, maintains a moderate register within each
voice, and gives both voices well-shaped trajectories with clear phrase-level
development and closure.
Table~\ref{tab:ladder-full} lists the exact per-model values. All four models
improve on both candidate checks with the harness, converging to a
$0.53$--$0.63$ collision and serialisation-consistency pass rate, a
$0.43$--$0.52$ constraint-checked delivery yield, and a $0.047$--$0.053$ degeneracy
band.

\begin{table}[h]
\centering
\caption{Per-model raw vs.\ harness results ($n{=}120$ per condition). Both
checks count unparseable raw candidates as failures; raw degeneracy is
conditional on parseability.}
\label{tab:ladder-full}
\small
\setlength{\tabcolsep}{2.5pt}
\resizebox{\columnwidth}{!}{%
\begin{tabular}{@{}lcccccc@{}}
\toprule
& \multicolumn{2}{c}{Ind.\ C+S} & \multicolumn{2}{c}{Checked yield}
& \multicolumn{2}{c}{Degeneracy} \\
Model & Raw & Harness & Raw & Harness & Raw & Harness \\
\midrule
DeepSeek V4 Flash & 0.350 & 0.558 & 0.125 & 0.433 & 0.186 & 0.053 \\
GPT-5 Nano        & 0.267 & 0.617 & 0.058 & 0.517 & 0.132 & 0.050 \\
DeepSeek V4 Pro   & 0.458 & 0.533 & 0.225 & 0.483 & 0.183 & 0.053 \\
Qwen3 235B        & 0.267 & 0.625 & 0.125 & 0.492 & 0.244 & 0.047 \\
\bottomrule
\end{tabular}
}
\end{table}

\paragraph{Cost details.}
The full harness uses about $157\times$ the raw token count and $11\times$ its
wall-clock time, costing approximately \$0.07 per Qwen3 235B piece. Skill
prompts account for most of the token overhead; estimates exclude local
verification, rendering, and storage.

\section{Expert Evaluation Protocol and Panel}
\label{app:expert}

\paragraph{Panel.}
Five expert raters evaluate the blind package. All are active composers and/or
music theorists with graduate training in composition and professional experience
in twelve-tone or atonal writing and analysis. Four raters have over twenty
years of professional practice; one has over thirty.

\paragraph{Protocol.}
Scores are generated once per condition with the same mid-tier model on eight
tasks drawn from the controlled task set, spanning $1$--$4$ voices and
$8$--$64$ bars (imitation, register contrast, rhythmic roles, and counterpoint
density). The study evaluates retained candidates irrespective of release-gate
status and is therefore a candidate-quality diagnostic. The preset comprises L1 counterpoint ($8$ bars), L2 imitation
($16$), L3 rhythmic roles ($32$), L4 register contrast ($64$), and V1--V4 at
$1$--$4$ voices respectively (counterpoint density, rhythmic roles, register
contrast, and imitation; all $8$ bars). Pair order, anonymous item codes, and
A/B assignment are pseudorandomly generated with a fixed seed. Each
task yields three anonymised pairwise comparisons: full harness versus raw LLM,
versus harness without skill injection, and versus harness without repair
($24$ pairs total). Within a pair the A/B assignment is randomised and
independent across pairs; one mapping is shared by all raters rather than
counterbalanced separately per rater, and the same full-harness candidate recurs in
the three contrasts for a task. The mapping to conditions is held in a separate
manifest not shown to raters. Raters receive English instruction sheets and two
printable scores per pair. The rating sheet enforces a forced binary choice (A or
B, no ties) on five dimensions defined below. Each rater completes all $24$
pairs.

\paragraph{Rating dimensions.}
Raters judge each pair on five forced-choice dimensions:
\begin{itemize}
\item \textbf{Adherence}: whether the score satisfies the \emph{specific}
task brief (voice count, bar length, imitation, register contrast, rhythmic
roles, etc.).
\item \textbf{Legality}: perceived twelve-tone rule conformance (row
order, avoidance of illegal octave doubling, no vertical pitch-class collisions).
\item \textbf{Style}: idiomatic serial writing (contour, register use,
avoidance of tonal implication, motivic interest).
\item \textbf{Coherence}: large-scale structure and sense of intentional
form.
\item \textbf{Overall}: global preference between the two scores.
\end{itemize}

\paragraph{Aggregation.}
For each (comparison, dimension) we count how often the full harness is chosen
($n=40$ repeated judgements: $5$ raters $\times$ $8$ tasks). We report the
descriptive win rate and a $10{,}000$-resample task-clustered bootstrap CI:
each draw resamples the eight tasks and retains all five ratings for each sampled
task (fixed seed $7$). We also report Fleiss' $\kappa$ across the five binary ratings per pair.
We do not treat the $40$ ratings as independent observations.

\begin{table}[t]
\centering
\caption{Expert pairwise results ($n=40$ repeated judgements per cell).
Win-rate $=$ fraction preferring the full harness.}
\label{tab:expert-full}
\setlength{\tabcolsep}{3.5pt}
\small
\begin{tabular}{@{}llcc@{}}
\toprule
Comparison & Dimension & W / L & Win-rate \\
\midrule
\multicolumn{4}{@{}l}{\emph{Full harness vs.\ raw generation}} \\
  & Adherence  & 34 / 6  & 0.85 \\
  & Legality   & 34 / 6  & 0.85 \\
  & Style      & 28 / 12 & 0.70 \\
  & Coherence  & 33 / 7  & 0.82 \\
  & Overall    & 36 / 4  & 0.90 \\
\midrule
\multicolumn{4}{@{}l}{\emph{Full harness vs.\ without skills}} \\
  & Adherence  & 24 / 16 & 0.60 \\
  & Legality   & 30 / 10 & 0.75 \\
  & Style      & 25 / 15 & 0.62 \\
  & Coherence  & 23 / 17 & 0.57 \\
  & Overall    & 22 / 18 & 0.55 \\
\midrule
\multicolumn{4}{@{}l}{\emph{Full harness vs.\ without repair}} \\
  & Adherence  & 30 / 10 & 0.75 \\
  & Legality   & 26 / 14 & 0.65 \\
  & Style      & 25 / 15 & 0.62 \\
  & Coherence  & 23 / 17 & 0.57 \\
  & Overall    & 24 / 16 & 0.60 \\
\bottomrule
\end{tabular}
\end{table}

\paragraph{Full results.}
Table~\ref{tab:expert-full} lists descriptive counts for all three comparisons.
For full versus raw, task-clustered $95\%$ CIs are $[.75,.95]$ (adherence),
$[.725,.95]$ (legality), $[.60,.775]$ (style), $[.75,.90]$ (coherence), and
$[.80,.975]$ (overall). Per-rater rates pooled over dimensions range from
$0.60$ to $1.00$. Fleiss' $\kappa$ ranges from $-.13$ to $.03$ across these
dimensions, indicating weak agreement despite the aggregate preference.
Across all three contrasts and dimensions, $\kappa$ ranges from $-.20$ to $.20$.

\paragraph{Reproducibility.}
The blind evaluation package ships with the release: anonymised score pairs,
English task sheets, rater forms, the hidden condition manifest, and aggregated
tally tables matching Table~\ref{tab:expert-full}.

\section{Recruitment and Payment}
\label{app:recruitment}

Participation is voluntary and uncompensated, consistent with specialist
academic consultation. The study collects only anonymised pairwise score
preferences and, given this design, no dedicated institutional ethics
review is sought.

\section{Reproducibility and Release Artefacts}
\label{app:repro}

We release the harness, $40$-task brief bank, frozen $20$-piece reference corpus, blind-evaluation package, and
scripts for computing the independent collision and serialisation-consistency
check, final constraint check, degeneracy, process-tier acceptance,
and corpus-distance results.

\section{Use of LLMs}
\label{app:llm_use}

LLMs have two distinct roles in this work. First, the models named in the
experimental design are evaluated systems: their generated proposals or raw
scores constitute experimental observations. Second, LLM-based assistants
support manuscript editing such as grammar, style, and clarity improvements.
We retain full control over the content, phrasing, and presentation of the final
text.

\end{document}